\documentclass[runningheads]{waica}

\usepackage{cmap}
\usepackage[T1]{fontenc}
\input{glyphtounicode}
\usepackage{lmodern}
\usepackage{graphicx}
\usepackage{booktabs}
\usepackage{multirow}
\usepackage{array}
\usepackage{amsmath}
\usepackage{amssymb}
\usepackage{xcolor}
\usepackage{url}
\usepackage{hyperref}

\newcommand{\pearl}{PEARL}
\newcommand{\gir}{\ensuremath{\mathcal{G}}}
\newcommand{\nodes}{\ensuremath{\mathcal{V}}}
\newcommand{\edges}{\ensuremath{\mathcal{E}}}

\newcommand{\cameraReadyAuthors}{%
Bohan Su\inst{1,*}\and
Pengze Li\inst{2,*}\and
Yuchen Lu\inst{3}\and
Xi Chen\inst{4,\dagger}}
\newcommand{\cameraReadyAuthorRunning}{B. Su et al.}
\newcommand{\cameraReadyInstitutes}{%
School of Computer Science, Wuhan University\\
\email{bohansu@whu.edu.cn}
\and
Artificial Intelligence Innovation and Incubation, Fudan University\\
\email{lipz25@m.fudan.edu.cn}
\and
Department of Computer Science, Faculty of Science, University of Bath\\
\email{yl4002@bath.ac.uk}
\and
College of Computer Science and Artificial Intelligence, Fudan University\\
\email{x\_chen@fudan.edu.cn}\\[0.5ex]
\textsuperscript{*}Equal contribution. \textsuperscript{$\dagger$}Corresponding author.}

\begin{document}

\title{\pearl{}: Auditable Repair for Scientific Reasoning Graph Extraction}
\titlerunning{\pearl{} for Scientific Reasoning Graph Extraction}

\author{\cameraReadyAuthors}
\authorrunning{\cameraReadyAuthorRunning}
\institute{\cameraReadyInstitutes}

\maketitle

\begin{abstract}
Scientific Reasoning Graph Extraction (SRGE) aims to recover explicit links among observations, evidence, intermediate claims, and paper-level conclusions.
LLMs can produce graph-like scientific explanations, but their outputs often mix malformed syntax, drifting edge labels, incorrectly oriented roots, and weak source anchors.
We propose \pearl{} (\emph{Peircean Extraction via Abstraction and Repair Layer}), a training-free framework that turns noisy LLM graph responses into auditable reasoning graphs and repairs them toward strict semantic validity.
\pearl{} first materializes explicit graph content under a closed Peircean schema, then uses matched evidence-grounded judge feedback to repair rejected edge types, local inference steps, and terminal roots while preserving an audit trail.
On five 70-paper model archives from ARCHE, a benchmark for latent reasoning-chain extraction, \pearl{} raises strict gate passes from $0/350$ for the LLM baseline to $300/350$, with average REA improving from $0.339$ to $0.906$.
The graphs provide a reliability layer for research-agent and AI scientist workflows that need inspectable reasoning traces rather than unconstrained graph regeneration.
Code and audit artifacts are available at \url{https://github.com/BohanSu/auditable-repair-reasoning-graphs/tree/300-350_workshop}

\end{abstract}

\section{Introduction}

Scientific agents increasingly need to read papers, retain evidence, compare claims, and reuse intermediate reasoning in later hypothesis generation and experiment planning.
For such research-agent workflows, a paper is not only a document to summarize. It is also a source of structured scientific reasoning.
Scientific Reasoning Graph Extraction (SRGE) targets this structure by recovering links among observations, cited evidence, intermediate claims, and paper-level conclusions~\cite{li2026arche,saha2021explagraphs}.
Recent LLMs can emit graph-like explanations with semantically useful node text and edge labels~\cite{saha2021explagraphs,li2026arche,mo2026kggen,zhang2025gkg}, but a graph-like response is not automatically a reliable graph artifact.
Raw responses can contain malformed DOT, extra prose, XML-like fragments, renamed edge labels, missing endpoints, multiple terminal roots, or roots oriented in the wrong direction.

This makes SRGE reliability a staged problem.
Structured generation and constrained decoding reduce surface-format failures~\cite{poesia2022synchromesh,geng2025jsonschemabench}. Scientific extraction and knowledge-graph systems populate structured fields or graph representations~\cite{li2025exploring,mo2026kggen,zhang2025gkg}. Verification benchmarks test whether claims are supported by evidence~\cite{javaji2025can,wang2025sciver}.
However, SRGE release forces all of these concerns into one artifact. The graph must be parseable, conform to a closed role taxonomy, preserve valid endpoints and terminal-root structure, and later satisfy semantic evaluation over entities and reasoning edges.
If parsing, repair, and evaluation are conflated, a pipeline may either reject salvageable graph material or silently regenerate a graph that no longer reflects the original model output.

\begin{figure}[t]
\centering
\includegraphics[width=\textwidth]{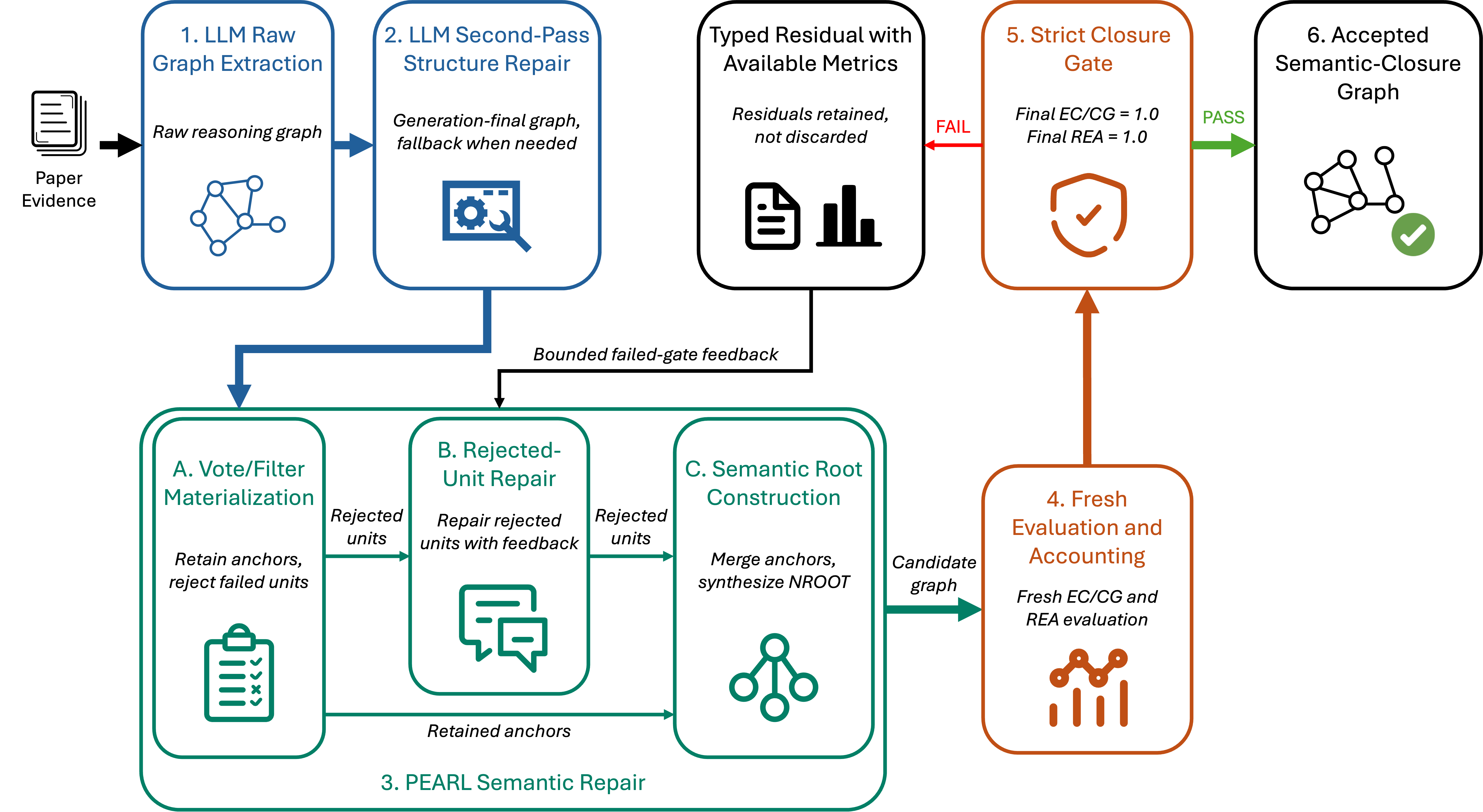}
\caption{Overview of the \pearl{} framework.}
\label{fig:framework}
\end{figure}

\pearl{} addresses this gap by treating raw LLM graph responses as conditional structural material for later repair rather than trusted graph artifacts.
The pipeline first materializes explicit response content under a fixed source-referenced graph contract and converts paper evidence into a generation-final graph through LLM extraction plus autonomous structure repair.
It then applies vote/filter materialization, rejected-unit repair, and claim-root construction under logged local edits.
A fresh evaluation under a fixed strict gate determines whether the output is accepted as semantically valid or retained as a typed residual with metric and audit records.
This staged flow is summarized in Fig.~\ref{fig:framework}.
The central boundary is simple. Structural recovery may normalize and repair graph form, but semantic repair must be tied to matched judge feedback and re-evaluated before acceptance.

\pearl{} makes three contributions.
First, it defines a training-free SRGE materialization framework that converts heterogeneous LLM graph responses into auditable graph artifacts under a closed Peircean schema.
Second, it adds judge-feedback repair for semantic correctness. Rejected local inference steps, edge types, and terminal roots are repaired through logged local operations and then judged again under the same CG/REA protocol.
Third, it evaluates the framework on ARCHE across five LLM archives, showing that \pearl{} substantially improves strict semantic validity and releases graph outputs that can support later reasoning-data construction and AI scientist workflows.

\section{Related Work}
\label{sec:related_work}

\paragraph{Scientific Reasoning Graphs.}
Scientific reasoning and graph-explanation tasks define schemas linking claims, evidence, and argumentative relations~\cite{saha2021explagraphs,li2026arche}.
They clarify target graph content, but not release readiness for graph-like LLM responses.
\pearl{} complements these tasks by asking when an emitted SRGE response can be safely repaired and admitted.

\paragraph{Structured Generation and Extraction.}
Constrained decoding, grammar-aware prompting, and schema checks reduce format errors in LLM outputs~\cite{poesia2022synchromesh,geng2025jsonschemabench}.
Scientific information extraction and graph-construction systems populate structured fields, knowledge graphs, or retrieval-oriented graph indices from scientific text~\cite{mo2026kggen,zhang2025gkg,li2025exploring}.
These methods provide important upstream capabilities, but SRGE requires graph-level invariants over a single artifact, including closed Peircean labels, valid endpoints, terminal-root conventions, and source-reference fields.
Parser compliance alone does not guarantee valid roles or root structure.

\paragraph{Self-Repair and Verification.}
Self-repair, tool-use, multi-agent feedback, and judge-based repair improve model outputs in many settings~\cite{schick2023toolformer,wu2024autogen,zheng2023judging,madaan2023self}.
Scientific verification benchmarks separately evaluate whether generated claims are supported or contradicted by evidence~\cite{javaji2025can,wang2025sciver}.
\pearl{} combines these ideas with a stricter artifact boundary. Repair is not unconstrained whole-graph regeneration, but evaluator-guided correction over a matched source graph with final CG/REA re-evaluation.
The accepted artifact must also remain evidence-linked after repair, not just locally plausible to evaluators.

\section{Method}
\label{sec:method}

\pearl{} is a training-free, schema-first framework that turns noisy LLM graph outputs into auditable artifacts and, when possible, semantically valid graphs.
Its core design decouples structural admissibility from semantic validity.
The pipeline first stabilizes source-linked structure and then corrects reasoning errors using matched-judge feedback.
This staged approach improves strict-gate passes without unconstrained whole-graph regeneration.

\subsection{Source-Referenced Graph Contract}
\label{subsec:problem}

Given a paper and input fields, \pearl{} targets a directed reasoning graph
\begin{equation}
    \gir = (p, \nodes, \edges, r),
\end{equation}
where $p$ identifies the paper instance, $\nodes$ is the set of graph nodes, $\edges$ is the set of typed edges, and $r$ is the terminal root of the released graph.
Each node $v_i=(q_i,a_i,t_i)$ contains a local node identifier $q_i$, a source-reference tuple $a_i=(x_i,y_i,z_i)$, and natural-language text $t_i$.
The integer coordinates identify a sentence ($x_i$), a viewpoint or cited reference within it ($y_i$), and an opinion within that reference ($z_i$); $(0,0,0)$ denotes implicit information.
The source tuple is a structural pointer into the input paper fields rather than a proof of semantic entailment.

Each edge $e_j=(u_j,v_j,\tau_j)$ points from source node $u_j$ to target node $v_j$ and carries a Peircean role label~\cite{peirce1934collected}:
\begin{equation}
    \tau_j \in \mathcal{T} =
    \{\text{ded-rule}, \text{ded-case}, \text{ind-common},
    \text{ind-case}, \text{abd-phen}, \text{abd-know}\}.
\end{equation}
Structural admissibility requires closed edge labels, known endpoints, no self-loops, and a single-root graph after structural repair.
Target nodes must also have incoming support that can be assigned to a single Peircean reasoning family after normalization, meaning that the incoming roles for the target instantiate one coherent deduction, induction, or abduction pattern rather than mixing incompatible role families.
We call one target node together with its valid paired incoming Peircean edges a reasoning unit.
These are structural conditions. Semantic correctness is established only by later evaluator judgments.

\subsection{Tolerance-Aware Structural Materialization}
\label{subsec:gir}

Stage~1 admits explicit structural material without inventing absent semantics.
The pipeline extracts graph-like regions from the raw response and then attempts strict DOT parsing before invoking a syntax-tolerant recovery path.
The tolerant path accepts DOT-like shapes, loose delimiters, trailing prose, and explicit XML/plist node-edge fields only when the corresponding nodes and edges are present in the response surface.
It does not create missing scientific nodes, synthesize omitted edges, or complete partially specified claims.

The recovered output is normalized into a canonical graph representation, so downstream stages never depend on which parser succeeded.
The parser branch remains only as audit metadata.
From the normalized graph, \pearl{} derives a strict structural artifact for storage and an evaluation view for scoring.
If single-root storage requires an engineering aggregation root, that root is treated as storage scaffolding and removed from the evaluation view, preventing storage constraints from being mistaken for scientific conclusions.

\subsection{Deterministic Schema Repair}
\label{subsec:repair}

The recovered graph may still contain schema corruption, including edge-label aliases, invalid endpoints, self-loops, duplicate nodes, mixed-family targets, isolated nodes, or root-orientation errors.
\pearl{} applies deterministic schema repair before semantic judgment.
This stage canonicalizes edge-label aliases into the closed Peircean taxonomy, de-duplicates graph elements, removes invalid endpoints and self-loops, repairs root orientation under a fixed priority rule, and records every operation.
When satisfying structural constraints would require unsupported inference, the pipeline records a typed failure instead of treating repair as semantic correction.

This repair is semantic-aware but not semantically unconstrained.
For example, when an evaluator later rejects a reasoning unit because its Peircean edge type is wrong, \pearl{} can repair the edge/root configuration under that judgment.
But such changes are routed through the judge-feedback stage and must be re-evaluated. They are not silently folded into structural parsing.

\subsection{Judge-Feedback Semantic Repair}
\label{subsec:semantic_repair}

Stage~2 starts from the released source graph and its matched paper evidence under the same evaluation protocol.
\pearl{} first runs a source evaluation to obtain content grounding (CG) and reasoning edge accuracy (REA).
It also obtains local judge decisions over reasoning units, which separate majority-correct reasoning anchors from evaluator-rejected units.
This matched evaluation is required because repair decisions may not consume stale votes from a different graph or from an earlier repair state.

\pearl{} then performs vote-guided repair.
Majority-correct source reasoning is retained as the graph backbone, while rejected reasoning units are repaired under matched local judge feedback and reintroduced as repaired reasoning units.
The method also recovers a claim root that expresses the paper-level research claim or proposal.
Repair is therefore local and judge-conditioned. It targets rejected edge types, reasoning units, and roots rather than regenerating the entire graph from scratch.

The repaired graph is evaluated again under the same protocol.
Previously accepted source reasoning may keep clean votes only when retained unchanged. Repaired reasoning, changed targets, and claim-root nodes must be judged again.
If a repaired unit or root still fails, \pearl{} records the failure and may continue with bounded regeneration or pruning.

\subsection{Final Gate and Provenance Audit}
\label{subsec:evidence_audit}

A candidate is strictly accepted only if it satisfies the unchanged final gate:
\begin{equation}
\mathrm{CG}_{\mathrm{final}} = 1.0
\quad \text{and} \quad
\mathrm{REA}_{\mathrm{final}} = 1.0.
\end{equation}
This gate defines semantic validity in the current framework.
Non-accepted samples receive typed residual labels rather than being dropped.
Some residuals occur before a terminal graph is produced, for example when there is no reliable source anchor for repair. Others occur after final evaluation, for example when the repaired graph still misses the metric gate.

Provenance is audited separately from the strict final gate.
For lexical source-span diagnostics, \pearl{} scores each node against its referenced paper span using the larger of two overlap scores, token F1 and node-to-span containment.
At paper level, the audit reports whether most nodes have sufficient resolved source overlap and whether default-source nodes remain limited.
These diagnostics do not change node text or edges and do not replace semantic evaluation.
We also report MiniCheck~\cite{tang2024minicheck} as an external source-grounding supplement. It checks graph-node claims against graph-identified source sentences, but it is not the strict acceptance gate.

\section{Experiments}
\label{sec:experiments}

\subsection{Experimental Setup}

\paragraph{Data and Models.}
We evaluate on five model archives from ARCHE~\cite{li2026arche}, a benchmark for latent reasoning-chain extraction: Claude Sonnet 4.5~\cite{anthropic2025sonnet45}, Gemini 3.1 Pro Preview~\cite{google2026gemini31propreview}, GPT-5.2~\cite{openai2025gpt52}, Grok 4.1 Thinking~\cite{xai2025grok41}, and Qwen3.5-397B-A17B~\cite{qwen2026qwen35}.
Each archive contains 70 outputs under the SRGE prompt protocol, for $5 \times 70 = 350$ rows.

\paragraph{Baseline and Metrics.}
The LLM baseline is the generation-final graph after raw extraction and autonomous second-pass structure repair/fallback, before \pearl{} semantic repair.
The primary metrics are CG and REA.
The strict gate pass criterion is $\mathrm{CG}=1.0$ and $\mathrm{REA}=1.0$.
MiniCheck source-grounding results are reported as diagnostics only.

\subsection{Main Results}

The five-model CG/REA results appear in Table~\ref{tab:main}.
The LLM baseline does not pass the strict gate on any of the 350 rows, while \pearl{} passes on 300 rows.
Average CG improves from $0.827$ to $0.896$, and average REA improves from $0.339$ to $0.906$.
The larger REA gain shows that most of the remaining gap in the baseline is not entity mention coverage, but reasoning-edge and root correctness.
This pattern matters for deployment because many baseline graphs are not empty or wholly unusable. They often retain enough local content to look plausible while still failing the global consistency conditions required for reliable reuse.
In that failure regime, a repair layer is most valuable when it preserves anchored node content and targets edge/root correctness directly instead of replacing the entire graph with unconstrained regeneration.

\begin{table}[t]
\centering
\small
\caption{Five-model SRGE results on ARCHE. Base is the LLM baseline before \pearl{} repair; \pearl{} is the terminal graph after judge-feedback repair.}
\label{tab:main}
\setlength{\tabcolsep}{2pt}
\begin{tabular}{>{\raggedright\arraybackslash}p{4.2cm}cccccc}
\toprule
\multirow{2}{*}{Model} & \multicolumn{2}{c}{Strict Passes} & \multicolumn{2}{c}{CG} & \multicolumn{2}{c}{REA} \\
\cmidrule(lr){2-3}\cmidrule(lr){4-5}\cmidrule(lr){6-7}
 & Base & \pearl{} & Base & \pearl{} & Base & \pearl{} \\
\midrule
Claude Sonnet 4.5~\cite{anthropic2025sonnet45} & 0/70 & 61/70 & .785 & .883 & .219 & .886 \\
Gemini 3.1 Pro Preview~\cite{google2026gemini31propreview} & 0/70 & 53/70 & .698 & .806 & .411 & .829 \\
GPT-5.2~\cite{openai2025gpt52} & 0/70 & 68/70 & .962 & .992 & .467 & 1.000 \\
Grok 4.1 Thinking~\cite{xai2025grok41} & 0/70 & 59/70 & .823 & .889 & .277 & .901 \\
Qwen3.5-397B-A17B~\cite{qwen2026qwen35} & 0/70 & 59/70 & .869 & .910 & .319 & .916 \\
\midrule
\textbf{Aggregate} & \textbf{0/350} & \textbf{300/350} & \textbf{.827} & \textbf{.896} & \textbf{.339} & \textbf{.906} \\
\bottomrule
\end{tabular}
\end{table}

\paragraph{Per-model behavior.}
\pearl{} improves REA for all five archives.
GPT-5.2 reaches the strict gate on 68 of 70 rows and has final average REA of 1.000.
The mixed-surface and open-weight archives are harder. Gemini, Grok, Claude, and Qwen retain typed residuals, but each still moves from zero baseline strict passes to at least 53 passes.
Across the 50 residual rows, the main typed causes are no reliable repair anchor (31 rows), final metric gate failure (13), metric regression (4), and final judge failure (2).
The residual breakdown therefore identifies a fairly specific bottleneck.
Most remaining failures arise before late-stage semantic acceptance, because the released source graph never provides a sufficiently trustworthy anchor for controlled repair.
By contrast, metric regression and final-judge failure are comparatively rare, which suggests that matched re-evaluation prevents most harmful over-repair once a usable anchor is available.

The residual mix shows that archive-level gaps reflect both upstream recoverability and downstream repair.
GPT-5.2 enters Stage~2 with cleaner source-linked reasoning units, whereas harder archives leave fragments with weak or incomplete source anchors.
A lower pass count therefore need not indicate repair failure on an otherwise stable source graph.
The residual taxonomy separates missing-anchor failures from graphs that reach re-evaluation but miss the metric gate under the same fixed evaluation protocol.

The hard rows also concentrate cost asymmetrically.
When a reliable anchor is absent, extra evaluator rounds tend to revisit the same unresolved evidence boundary instead of producing new admissible structure.
This pattern suggests that future efficiency gains are more likely to come from stronger source-link recovery before repair than from simply extending the number of semantic re-judgment cycles after repair has already stalled.

\subsection{Diagnostic Ablation}
\label{subsec:ablation}

We ablate the terminal graph used for final evaluation over the same five-model, 350-row scope.
Raw extraction and the generation-final LLM graph nearly never pass the strict gate, with $1/350$ and $0/350$ passes.
No root gen. omits semantic-root construction, Retained root keeps the original root without rejected-unit repair, and Root only performs root repair without iterative feedback.
They reach $130/350$, $125/350$, and $102/350$ passes, while full \pearl{} reaches $300/350$.
The pattern shows that deterministic retention alone leaves many graphs open and root-only repair leaves evaluator-rejected local reasoning unresolved.
Reliable paper-level closure depends on stable local units first, so a stronger root alone cannot compensate for reasoning edges that still fail re-evaluation.
After local repair under matched judgments, the final claim root has a more stable support base.
The gap between Root only and full \pearl{} further shows that local reasoning errors propagate upward, so claim-root repair cannot rescue a graph whose supporting edge families remain inconsistent.

\subsection{Source-Grounding Supplement}
\label{subsec:minicheck}

MiniCheck evaluates whether graph-node claims are supported by the graph-identified source sentences.
Across raw extraction, the LLM second-pass graph, and the \pearl{} terminal graph, the scoreable rows are $345/350$, $350/350$, and $317/350$; support rates are $.511$, $.584$, and $.557$; and no-evidence rates are $.267$, $.141$, and $.184$.
These numbers are a source-grounding supplement rather than a replacement for CG/REA, because the stages have different node sets and do not define one monotonic score.
We use MiniCheck as an audit aid for source resolution quality, not as permission to bypass the stricter graph-level acceptance rule.
Rows that fail the strict gate can still preserve locally grounded node text for audit, while rows that pass after repair may still deserve source-link review before downstream reuse.
The drop from $350$ to $317$ scoreable terminal rows also shows that strict graph acceptance and node-level source checking fail for different reasons.
This split gives downstream users a conservative signal for deciding when source links need manual audit.
It also prevents local grounding from masking unresolved reasoning-edge errors.

\section{Limitations}

\pearl{} targets the ARCHE Peircean SRGE schema and five text-only archives. Portability to new schemas or multimodal evidence requires revised validators, frozen judge versions, and wider-deployment evidence checks.

\section{Conclusion}

\pearl{} provides auditable repair for LLM-generated scientific reasoning graphs under a strict CG and REA gate.
Across five ARCHE archives, it raises strict gate passes from $0/350$ to $300/350$, showing that graph-like LLM outputs need audited repair before final acceptance in practical scientific extraction workflows.

\bibliographystyle{waica}
\begin{credits}
\subsubsection{\discintname}
The authors have no competing interests to declare that are relevant to the content of this article.
\end{credits}

\bibliography{custom}

\end{document}